\documentclass[11pt]{article}

\usepackage[preprint]{acl}

\usepackage{times}
\usepackage{latexsym}
\usepackage{xcolor}
\usepackage{tcolorbox}
\usepackage[T1]{fontenc}
\usepackage[utf8]{inputenc}

\usepackage{microtype}

\usepackage{inconsolata}

\usepackage{graphicx}

\usepackage{tcolorbox}
\tcbuselibrary{listings,breakable,skins}
\usepackage{listings}

\lstdefinelanguage{json}{
  basicstyle=\ttfamily\footnotesize,
  morestring=[b]",
  stringstyle=\ttfamily\footnotesize,
  showstringspaces=false
}

\title{Beyond Perplexity: A Lightweight Benchmark for Knowledge Retention in Supervised Fine-Tuning}

\author{
  Soheil Zibakhsh\thanks{Equal contribution.}%
   \quad
  Pedram Aghazadeh\footnotemark[\value{footnote}] \quad
  Farinaz Koushanfar \\
  University of California San Diego \\
  \texttt{\{szibakhshshabgahi,paghazadeh,farinaz\}@ucsd.edu}
}

\begin{document}
\maketitle

\vspace{-2cm}
\begin{abstract}
Supervised Fine-Tuning (SFT) is a standard approach for injecting domain knowledge into Large Language Models (LLMs). However, relying on validation perplexity to monitor training is often insufficient, as it confounds stylistic mimicry with genuine factual internalization. To address this, we introduce the Knowledge Retention (KR) Test , a lightweight, corpus-grounded evaluation framework designed to distinguish factual learning from linguistics. KR-Test utilizes automatically generated contrastive examples to measure likelihood preferences for correct versus incorrect continuations, requiring no instruction tuning or generative decoding. We validate the framework’s integrity through a "blind vs. oracle" baseline analysis. Furthermore, we demonstrate the diagnostic capabilities of KR-Test by analyzing the training dynamics of Low-Rank Adaptation (LoRA). By exposing the fine-grained dissociation between linguistic convergence and knowledge retention, KR-Test enhances the interpretability of fine-tuning dynamics.
\end{abstract}

% --------------------
\section{Introduction}
% --------------------

% We validate the integrity of KR-Test using an oracle model with direct access to the source text, establishing a near-ideal upper bound on achievable performance.
% We then demonstrate the utility of KR-Test through an analysis of parameter-efficient fine-tuning, showing how different LoRA configurations and model capacities lead to markedly different levels of factual retention—differences that are not apparent from perplexity alone.

%%%%%%%%%%%%%%%%%%%%%%%%%%%%%%%%%

Large Language Models (LLMs) underpin a wide range of modern applications, from creative generation to decision support systems~\cite{achiam2023gpt, touvron2023llama}.
However, in many applied settings, such as legal analysis, scientific assistance, or domain-specific question answering, model utility depends not only on linguistic fluency, but on the faithful internalization of factual knowledge~\cite{zhang2025siren}.
A common strategy for injecting domain knowledge is SFT on curated corpora, often followed by instruction tuning~\cite{ouyang2022training, wei2021finetuned}.

During SFT, training progress is typically monitored using validation perplexity. While perplexity is an effective performance indicator, it aggregates token-level prediction errors and does not explicitly distinguish between stylistic learning and factual knowledge. Consequently, a model may achieve state-of-the-art perplexity by mimicking the stylistic contours of a dataset while failing to internalize the underlying knowledge, or worse, hallucinating plausible but incorrect facts~\cite{ji2023survey}. Therefore, having a direct signal for tracking whether a model has internalized the facts contained in the training data can be extremely useful.

Downstream Question Answering (QA) benchmarks offer one avenue for evaluating factual knowledge, but they are expensive to run frequently and often require instruction-following capabilities that are absent during early or intermediate stages of SFT.
This creates a practical gap: there is no lightweight, corpus-grounded evaluation useful for monitoring factual learning throughout training.

We address this gap by introducing the \emph{Knowledge Retention (KR) Test}, a likelihood-based evaluation framework designed to measure factual consistency with respect to the training corpus itself.
KR-Test automatically generates contrastive examples consisting of a shared context and two plausible continuations, one factually correct and one incorrect, and evaluates models by comparing conditional likelihoods. The context does not contain any hints for the correct continuation, rather it grounds the model into the passage that the facts come from.
The test requires no instruction tuning, avoids generative decoding, and can be integrated into standard validation pipelines with minimal overhead.

% Have we connected blind & oracle?
We validate the integrity of KR-Test through a "blind vs. oracle" baseline analysis, proving that our discriminative tasks are both non-trivial for pre-trained models and solvable given the context. Furthermore, we utilize KR-Test to analyze the training dynamics of Low-Rank Adaptation (LoRA) \cite{hu2022lora}, revealing a difference in learning capabilities when applying Parameter Efficient Fine-Tuning (PEFT) on different modules.

\begin{figure*}[t]
    \centering
    \includegraphics[width=1\textwidth]{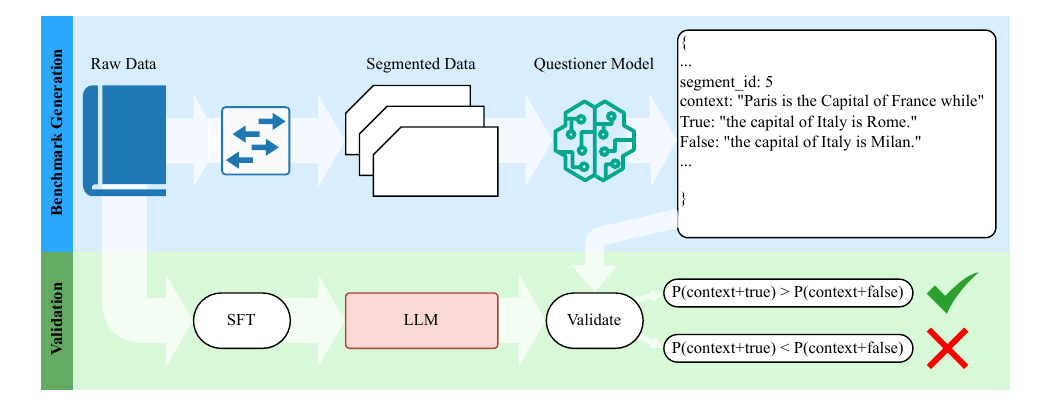}
    \caption{KR-Test generation and validation pipeline. Questions are generated using segmented data and are used in validation to track the model's learning progress.}
    \label{fig:pipeline}
\end{figure*}

\paragraph{Contributions.}
\begin{itemize}
    \item We introduce \textbf{KR-Test}, a lightweight, corpus-grounded evaluation framework for measuring factual retention during SFT.
    \item We validate the soundness of the benchmark via an oracle-based upper bound.
    \item We demonstrate how KR-Test provides novel insights into PEFT dynamics, including LoRA module placement and model capacity effects.
\end{itemize}

\section{Limitations of Perplexity for Factual Evaluation} % Instead of Why Perplexity is Insufficient
% --------------------

Perplexity measures average next-token predictive fit to a validation corpus and is therefore dominated by frequent tokens and local syntactic regularities.
Factual retention, however, concerns whether specific, often low-frequency facts from the training data are encoded such that the model prefers correct continuations over plausible but incorrect ones.
As a result, changes in factual internalization may have little effect on perplexity.
This motivates a complementary, corpus-grounded evaluation that directly probes factual consistency, which we address with KR-Test.

% --------------------
\section{Knowledge Retention Test}
% --------------------
The KR-Test is a lightweight validation framework for measuring factual retention during SFT.
Unlike static benchmarks, KR-Test is dynamically derived from the training corpus and evaluates whether a model prefers factually correct continuations over plausible but incorrect alternatives under identical contexts.
As illustrated in Figure~\ref{fig:pipeline}, KR-Test consists of three stages: segmentation, contrastive generation, and likelihood evaluation.

\paragraph{Semantic Segmentation.} To address unstructured domain corpora, we employ a high-capacity ``Teacher'' LLM to decompose raw text into discrete, disjoint passages. The Teacher is conditioned to ensure each passage is self-contained, encapsulating an atomic unit of information independent of the surrounding context.

\paragraph{Contrastive Generation.} The Teacher distills specific facts into sets of discriminative tasks. For each passage, it generates $N$ binary contrastive tuples $\tau = (x_c, x^+, x^-)$, consisting of a context, a factually correct continuation, and a plausible incorrect one. 

To ensure benchmark utility and remove stylistic artifacts, we enforce two critical constraints:
\begin{enumerate}
    \item \textbf{Correctness:} The verification of $x^+$ must be strictly grounded in the source passage.
    \item \textbf{Adversarial Similarity:} We explicitly constrain the Teacher to generate $x^-$ with similar length, syntax, and style to $x^+$. The continuations must differ only in factual content, preventing the model from utilizing length heuristics or surface-level priors.
\end{enumerate}
Prompt templates are detailed in Appendix~\ref{appx:curation}.

\paragraph{Likelihood Evaluation.} KR-Test evaluates a model using conditional likelihood, avoiding generative decoding.
To eliminate sequence-length bias, we compare cumulative log-probabilities up to the shorter continuation length $T = \min(|x^+|, |x^-|)$ and deem a sample correct if:
\begin{equation}
\sum_{t=1}^{T} \log p(x^+_t \mid x_c, x^+_{<t}) >
\sum_{t=1}^{T} \log p(x^-_t \mid x_c, x^-_{<t}).
\end{equation}
This requires only two forward passes per example, making KR-Test computationally comparable to standard perplexity validation.

% If we have space 

\subsection{Test Curation}

KR-Test is fully open-sourced\footnote{\url{https://github.com/soheilzi/KR-Test}} and can be applied to arbitrary domain corpora.
Questions are generated by an oracle (e.g., human annotators or frontier language models), with all curation safeguards and filtering criteria detailed in Appendix~\ref{appx:curation}.

\vspace{0.5em}
\noindent
An example KR-Test instance is shown below.

\begin{tcolorbox}[
  title={Example KR-Test instance},
  colback=gray!5,
  colframe=gray!40,
  breakable
]
\textbf{Context:}  
Geopyxis carbonaria has been reported for the first time from Turkey in 2010.

\vspace{0.5em}
\textbf{\textcolor{green!60!black}{Factually Correct Continuation (True):}}  The North American distribution of this fungus extends north to Alaska.

\vspace{0.3em}
\textbf{\textcolor{red!70!black}{Factually Incorrect Continuation (False):}}  
The North American distribution of this fungus extends only to the southern United States.

\vspace{0.5em}
\textbf{Source Location:} Passage~944 in the WikiText2 training corpus.
\end{tcolorbox}
% --------------------
\section{Experiments}
% --------------------
\subsection{Oracle-Based Validation}
\label{subsec:oracle_validation}

To validate the soundness of KR-Test, we estimate an empirical upper bound using an oracle model with direct access to the source paragraph from which each question is derived.
We construct this \emph{golden standard} on WikiText2~\cite{merity2016pointer}, providing a near-ideal reference that isolates evaluation quality from model limitations.

% \refstepcounter{figure}
% \begin{tcolorbox}[title={KR-Test oracle prompt}, breakable]
% \begin{lstlisting}
% *** Reference Paragraph ***
% {paragraph_text}

% *** Questions ***
% [Question ID: 1]
% Context: {context_1}
% Option A: {option_a}
% Option B: {option_b}

% ...

% [Question ID: n]
% Context: {context_n}
% Option A: {option_a}
% Option B: {option_b}
% \end{lstlisting}
% \end{tcolorbox}

% \captionof{figure}{Oracle prompt used for golden-standard KR-Test evaluation.}
% \label{fig:oracle_prompt_main}

We use the OpenAI \textbf{gpt-4o-mini}~\citep{achiam2023gpt} and supply the oracle with the reference paragraph and all KR-Test questions derived from it.
For each question, the oracle selects the factually consistent continuation based solely on the provided paragraph.
The oracle achieves \textbf{99.56\%} accuracy, indicating that the generated questions are largely unambiguous and that KR-Test faithfully reflects factual consistency. Conversely, standard pre-trained models perform near a random baseline demonstrated in Fig.~\ref{fig:llama}, confirming that the tasks are non-trivial and cannot be resolved through surface-level cues alone.
Complete prompts, decoding parameters, and additional analysis are provided in Appendix~\ref{appx:golden_standard}.

% --------------------
% --------------------
\subsection{Demystifying PEFT Dynamics with KR-Test}
\label{subsec:experiments}

% Parameter-Efficient Fine-Tuning (PEFT) has emerged as the standard paradigm for adapting Large Language Models (LLMs) to downstream tasks under constrained computational budgets. Among these techniques, Low-Rank Adaptation (LoRA) \cite{hu2022lora} is ubiquitous, allowing for the update of dense layers via low-rank decomposition matrices while keeping the pre-trained weights frozen.

PEFT is the dominant approach for adapting large language models under constrained budgets. Among these methods, Low-Rank Adaptation (LoRA)~\cite{hu2022lora} enables efficient updates by inserting low-rank adapters while keeping pre-trained weights frozen.

Despite its widespread adoption, it remains unclear which transformer components most effectively internalize \emph{new factual knowledge} during adaptation. While increasing the LoRA rank generally improves modeling capacity, standard metrics such as perplexity are insufficient to distinguish stylistic adaptation from factual acquisition.

% Despite its widespread adoption, the optimal allocation of trainable parameters within the transformer architecture remains an open research question. While increasing the rank $r$ generally correlates with improved modeling capability, it is unclear which specific architectural components—Attention mechanisms or Feed-Forward Networks (FFNs)—are most efficient at internalizing \textit{new factual knowledge} from a domain-specific corpus. Standard evaluations often rely on perplexity, which, as established in Section \ref{sec:intro}, fails to distinguish between stylistic mimicry and factual acquisition.

Here, we use KR-Test to probe knowledge retention under different LoRA configurations. By isolating adapter placement, we show that KR-Test provides a fine-grained signal for comparing PEFT design choices that are otherwise indistinguishable under perplexity-based evaluation as shown in Figure~\ref{fig:lora_efficiency}.

% In this section, we utilize the KR-Test to systematically evaluate the relationship between parameter placement and knowledge retention. By isolating specific modules, we demonstrate that KR-Test serves as a high-resolution probe for identifying optimal training configurations.

\subsubsection{Setup}
We fine-tune \texttt{Llama-3.2-1B} on WikiText2~\cite{merity2016pointer} for 5{,}000 steps under a fixed parameter budget.
We compare LoRA adapters applied to \textbf{Attention} layers (query, key, value, output) versus \textbf{Feed-Forward Networks (FFN}) blocks (gate, up, down), and evaluate factual retention using the KR-Test.

\subsubsection{Results and Analysis}
\begin{figure}[t]
    \centering
    \includegraphics[]{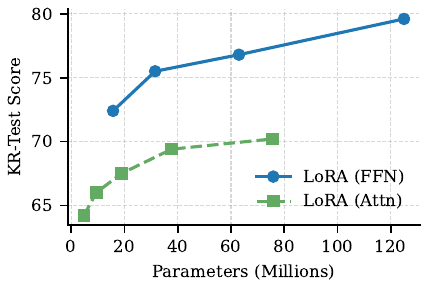}
    \caption{Parameter efficiency of LoRA module placement according to KR-Test.}
    \label{fig:lora_efficiency}
\end{figure}

% As illustrated in Figure \ref{fig:lora_efficiency}, we observe a significant discrepancy in learning efficiency between the two configurations. Given an identical parameter budget, LoRA modules placed on the \textbf{FFN} layers exhibit substantially higher knowledge retention scores compared to those placed on Attention layers.
Figure~\ref{fig:lora_efficiency} shows a clear efficiency gap between configurations.
Under identical parameter budgets, LoRA adapters placed on \textbf{FFN} layers consistently achieve higher KR-Test scores than those placed on Attention layers.

% This empirical finding aligns with recent interpretability literature suggesting that FFN layers in transformer models act as key-value memories that store factual associations \cite{geva2021transformer, meng2022locating}, while attention mechanisms primarily manage token routing and contextual dependencies. 

This result is consistent with prior findings that FFN layers encode factual associations, while Attention primarily supports token routing~\cite{geva2021transformer, meng2022locating}.

% Crucially, standard perplexity metrics often fail to capture this distinction, showing similar convergence rates for both configurations due to the model's ability to learn local syntax (style) regardless of module placement. The KR-Test, however, reveals that for tasks requiring strict adherence to source knowledge, the FFN layers are the more efficient vessel for adaptation in the \texttt{Llama-3.2-1B} architecture.

\subsubsection{Implications for Model Selection}
% We note that while the superiority of FFN-targeted adaptation is pronounced for \texttt{Llama-3.2-1B}, we do not claim this holds universally across all architectures or model sizes. Rather, this experiment underscores the utility of the KR-Test as a diagnostic tool. By providing a lightweight, disentangled signal for knowledge acquisition, KR-Test empowers practitioners to empirically determine the optimal PEFT configuration for their specific model and data regime, replacing heuristic choices with data-driven architectural decisions.

While FFN-targeted adaptation is most pronounced for \texttt{Llama-3.2-1B}, we do not claim universality across architectures or scales.
Rather, this result highlights KR-Test as a lightweight diagnostic: it provides a disentangled signal for factual acquisition that enables empirical selection of PEFT configurations for a given model and data regime, beyond heuristic choices.

\subsection{Capacity and Knowledge Scaling}
\label{subsec:scaling}

% Finally, we apply the KR-Test to investigate the impact of base model scale on knowledge acquisition. We compare the training dynamics of the \texttt{Llama-3.2-1B} model against its larger counterpart, \texttt{Llama-3.1-8B}, under identical SFT conditions.

% As demonstrated in Figure~\ref{fig:llama}, the performance gap is twofold. First, the larger model exhibits a higher initial KR-score (intercept), reflecting a superior reservoir of prior knowledge embedded during pre-training. Second, the larger model achieves a significantly higher final convergence score. This divergence indicates that while parameter placement (FFN vs. Attention) optimizes efficiency, the absolute capacity to internalize and retrieve specific facts is governed by the total parameter count. These results extend traditional neural scaling laws \cite{kaplan2020scaling} from general loss minimization to the specific domain of factual retention, suggesting that ``memorization capacity" scales predictably with model size.

\begin{figure}[t]
    \centering
    \includegraphics[]{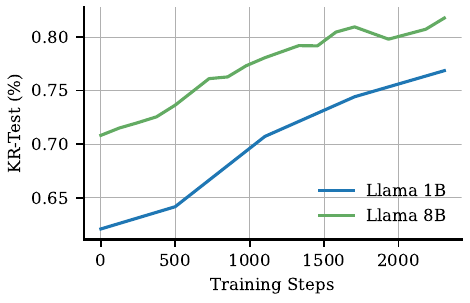}
    \caption{Effect of parameter count on model's initial and final knowledge retention}
    \label{fig:llama}
\end{figure}

We next examine the effect of base model scale using KR-Test, comparing \texttt{Llama-3.2-1B} with \texttt{Llama-3.1-8B} under identical SFT settings.
As shown in Figure~\ref{fig:llama}, larger models exhibit both higher initial KR scores, reflecting greater prior knowledge, and higher final convergence.

These results indicate that while adapter placement optimizes parameter efficiency, the absolute capacity for factual retention is primarily governed by model scale.
This extends neural scaling observations~\cite{kaplan2020scaling} to factual retention, suggesting that memorization capacity increases predictably with parameter count.

% --------------------
% \subsection{Knowledge Retention in RAG Systems}
% % --------------------

% We extend KR-test to Retrieval-Augmented Generation by conditioning likelihoods on retrieved documents.
% We evaluate the same KR questions in two modes:
% (i) parametric-only inference and
% (ii) retrieval-augmented inference.

% Our results reveal three regimes:
% (1) high KR without RAG but low QA accuracy, indicating poor answer extraction,
% (2) low KR but high RAG QA accuracy, indicating reliance on retrieval without internalization,
% and (3) failures where models assign high likelihood to retrieval-consistent but factually incorrect continuations.

% KR-test thus provides a diagnostic tool that disentangles parametric knowledge from retrieval-dependent behavior.

% --------------------
\section{Related Work}
% --------------------

\textbf{Factual Probing.}
Benchmarks such as LAMA \citep{petroni2019language} and its variants probe parametric knowledge using cloze-style queries, but are sensitive to prompting and largely disconnected from training corpora.

\textbf{Question Answering Benchmarks.}
QA datasets such as Natural Questions \citep{kwiatkowski2019natural}, PopQA \citet{mallen2022not}, and MMLU \citet{hendrycks2020measuring} evaluate answer correctness but conflate knowledge with retrieval, decoding, and reasoning strategies. Furthermore, these datasets are expensive to collect and to use frequently during training.

\textbf{Decoding and RL for Knowledge Extraction.}
Recent advances in self-consistency \citep{wang2022self}, tree-based decoding \citep{yao2023tree}, and RL for reasoning show that performance gains often arise from better utilization of existing knowledge rather than new learning \citep{zelikman2022star}.
KR-test complements these approaches by measuring whether knowledge exists to be extracted.

% --------------------
\section{Conclusion and Future Work}
% --------------------

By disentangling factual acquisition from stylistic alignment, KR-Test offers a granular view of SFT dynamics often unseen by perplexity. This separation enables the formulation of Knowledge Scaling Laws. Unlike standard laws based on aggregate loss \cite{kaplan2020scaling}, KR-Test isolates the "bit-rate" of factual learning; we hypothesize that knowledge retention follows a distinct trajectory where small models may saturate on facts despite continuing to improve on syntax. Additionally, future work must characterize the transfer function between discriminative preference and generative fidelity. Establishing the correlation between KR-Test likelihoods and downstream benchmarks (e.g., MMLU) is crucial to validate the metric as a compute-efficient proxy for early stopping and hyperparameter selection.
\newpage
\section{Limitations}

Our work has three primary limitations. First, KR-Test measures discriminative preference, not generative fidelity. While a higher likelihood for the correct continuation indicates knowledge internalization, it does not guarantee that the model will output the correct fact during unconstrained generation, where other hallucinations might occupy significant probability mass. Second, our analysis of parameter efficiency (Figure~\ref{fig:lora_efficiency}) serves to demonstrate the diagnostic granularity of KR-Test using Llama-3.2-1B. We do not claim that the superiority of FFN-targeted adaptation is universal across all model scales or architectures, but it would be interesting to analyze other families of models for such trends. Finally, the robustness of KR-Test relies on the quality of the automatically generated contrastive examples. While we validate the task difficulty via our "Blind vs. Oracle" baseline, semantic ambiguity or overly simple distractors in the evaluation set could potentially inflate retention scores.

\bibliographystyle{acl_natbib}
\bibliography{custom}
\newpage
\thispagestyle{empty}
\null
\newpage
\appendix
\setcounter{figure}{0}
\renewcommand{\thefigure}{A\arabic{figure}}
\section{Curation}\label{appx:curation}
Figure~\ref{fig:kr_example} shows an illustrative example of a single KR-Test instance derived from WikiText. Each instance consists of a short context and two candidate continuations that are syntactically similar but differ in factual correctness.

% \refstepcounter{figure}
\begin{tcolorbox}[title={KR-Test Example Instance}, breakable]
\begin{lstlisting}[language=json]
{
  "context": "The name 'Oldham' is believed to date from 865, during the period of the Danelaw.",
  "sentence_true": "The town's name is thought to derive from the Old Norse name Aldehulme.",
  "sentence_false": "The town's name is thought to derive from the Old Norse name Aldenhume.",
  "location": 13630
}
\end{lstlisting}
\end{tcolorbox}

\captionof{figure}{Example KR-Test instance derived from WikiText illustrating minimal factual perturbation.}
\label{fig:kr_example}

\paragraph{Field Definitions.}
Each KR-Test instance contains the following fields:

\begin{itemize}
    \item \texttt{context}: A short prefix extracted or paraphrased from the source paragraph that establishes the factual setting. The context is designed to be insufficient on its own to trivially resolve the question without internalized knowledge.
    
    \item \texttt{sentence\_true}: A factually correct continuation that is directly supported by the source text. When appended to the context, the resulting sequence reflects a truthful statement present in the training corpus.
    
    \item \texttt{sentence\_false}: A plausible but factually incorrect alternative continuation. This sentence is constructed to closely match the true continuation in length, syntax, and style, differing only in a key factual detail (e.g., entity name, relation, or attribute).
    
    \item \texttt{location}: The character offset of the source paragraph within the original corpus. This field is used for traceability and debugging but is not used during evaluation.
\end{itemize}

\paragraph{Evaluation Protocol.}
During evaluation, a model is presented with the same \texttt{context} and scored based on its conditional likelihood of \texttt{sentence\_true} versus \texttt{sentence\_false}. A test instance is considered correct if the model assigns higher likelihood to the factually consistent continuation under identical conditioning.

\section{Golden Standard}\label{appx:golden_standard}
This appendix documents the oracle-based \emph{golden standard} evaluation used to validate the soundness of the KR-Test.
The purpose of this evaluation is to estimate an empirical upper bound on KR-Test accuracy under near-ideal conditions, ensuring that errors observed in downstream models primarily reflect limitations in factual retention rather than artifacts of question construction.

\subsection{Source Paragraphs and Alignment}

Oracle evaluation is conducted on paragraphs extracted from WikiText2~\cite{merity2016pointer}, which serves as both the SFT corpus and the source of KR-Test questions.
Each KR-Test instance contains a \texttt{location} field indicating the starting line index of the source paragraph in the WikiText2 training split.
This index is used to deterministically retrieve the exact paragraph associated with each group of questions during oracle evaluation.

\subsection{Batched Oracle Evaluation}

KR-Test questions are evaluated in batches grouped by their shared \texttt{location}.
Each batch corresponds to a single source paragraph and includes all questions derived from that paragraph.
For each batch, the oracle is provided with the full reference paragraph and the complete set of associated questions, each consisting of a short context and two candidate continuations (Option A and Option B).
This batched formulation ensures that the oracle has access to all relevant local evidence while minimizing evaluation noise.

\subsection{Option Randomization and Decoding Constraints}

To prevent position-based bias, the ordering of the factually correct and incorrect continuations is randomized independently for each question.
The oracle is instructed to select the option that is factually consistent with the reference paragraph and to output responses in a strict JSON format mapping question identifiers to choices (\texttt{"A"} or \texttt{"B"}).
Decoding is performed deterministically (temperature set to zero) to ensure reproducibility.

\onecolumn
\subsection{Oracle Prompt (Example)}
\begin{lstlisting}[language=json]
[
  {"role":"system","content":"Instruction: You are an expert reading comprehension system. You will be provided with one Reference Paragraph and a list of Questions related to it. For each Question ID, determine which Option (A or B) is FACTUALLY CONSISTENT with the Reference Paragraph. "Output strictly in JSON format where keys are Question IDs and values are the choice strings 'A' or 'B'."},
  {"role":"user",
  "content":"*** Reference Paragraph ***
 The common starling ( Sturnus vulgaris ) , also known as the European starling , or in the British Isles just the starling , is a medium @-@ sized passerine bird in the starling family , Sturnidae . It is about 20 cm ( 8 in ) long and has glossy black plumage with a metallic sheen , which is speckled with white at some times of year . The legs are pink and the bill ...

*** Questions ***
        
            [Question ID: 1]
            Context: The common starling is a medium-sized passerine bird with glossy black plumage and a metallic sheen.
            Option A: Its legs are blue and the bill is red in summer.
            Option B: Its legs are pink and the bill is yellow in summer.
            
            [Question ID: 2]
            Context: The common starling has been introduced to various countries including Australia and New Zealand.
            Option A: It was introduced to Australia in 1857 to control insect pests.
            Option B: It was introduced to Australia in 1907 to control insect pests.
            
            [Question ID: 3]
            Context: Common starlings construct untidy nests in natural or artificial cavities.
            Option A: They lay four or five glossy, pale blue eggs in their nests.
            Option B: They lay six or seven glossy, pale blue eggs in their nests.
            
            [Question ID: 4]
            Context: The common starling's vocal repertoire is highly variable and includes sounds mimicked from other birds.
            Option A: Proficient male starlings can have a repertoire of up to 25 variable song types.
            Option B: Proficient male starlings can have a repertoire of up to 35 variable song types.
            
            [Question ID: 5]
            Context: The species is classified as least concern by the International Union for Conservation of Nature.
            Option A: Despite population increases in northern and western Europe, its global numbers are declining.
            Option B: Despite population declines in northern and western Europe, its global numbers are stable.
            
            [Question ID: 6]
            Context: Common starlings prefer urban or suburban areas for nesting and roosting.
            Option A: They are often found in grassy areas like farmland and golf courses for foraging.
            Option B: They are often found in dense, wet forests for foraging.
            
            [Question ID: 7]
            Context: The common starling is largely insectivorous but will also eat seeds and fruits when available.
            Option A: They primarily feed on insects such as beetles and grasshoppers.
            Option B: They primarily feed on fish such as salmon and mackerel.
            
            [Question ID: 8]
            Context: Common starlings can form very large flocks, creating murmurations that are a sight to behold.
            Option A: These flocks may consist of no more than a thousand individuals in some cases.
            Option B: These flocks may consist of over a million individuals in some cases.
]
\end{lstlisting}
\twocolumn

\end{document}